\documentclass[conference]{IEEEtran}
\usepackage{times}
\usepackage{epsfig}
\usepackage{graphicx}
\usepackage{amsmath}
\usepackage{amssymb}
\usepackage{booktabs}
\usepackage{multirow}
\usepackage{hyperref}
\usepackage{lipsum} 
\usepackage{float}  

\IEEEoverridecommandlockouts 

\begin{document}

\title{The 1st International Workshop on Disentangled Representation Learning for Controllable Generation (DRL4Real): Methods and Results\thanks{\textsuperscript{\dag}Workshop Organizers. \textsuperscript{*}Corresponding author: jinxin@eitech.edu.cn. \\
The DRL4Real website: \url{https://drl-for-real.github.io/DRL-for-Real/index.html} \\
The Competition website: \url{https://eval.ai/web/challenges/challenge-page/2527/overview}}}

\author{
\IEEEauthorblockN{
\small 
\begin{tabular}{@{}c@{\hspace{0.5em}}c@{\hspace{0.5em}}c@{\hspace{0.5em}}c@{\hspace{0.5em}}c@{\hspace{0.5em}}c@{}} 
Qiuyu Chen\textsuperscript{\dag} & Xin Jin\textsuperscript{\dag,*} & Yue Song\textsuperscript{\dag} & Xihui Liu\textsuperscript{\dag} & Shuai Yang\textsuperscript{\dag} & Tao Yang\textsuperscript{\dag} \\
Ziqiang Li\textsuperscript{\dag} & Jianguo Huang\textsuperscript{\dag} & Yuntao Wei\textsuperscript{\dag} & Ba'ao Xie\textsuperscript{\dag} & Nicu Sebe\textsuperscript{\dag} & Wenjun (Kevin) Zeng\textsuperscript{\dag} \\[0.8em]
Jooyeol Yun & Davide Abati & Mohamed Omran & Jaegul Choo & Amir Habibian & Auke Wiggers \\
Masato Kobayashi & Ning Ding & Toru Tamaki & Marzieh Gheisari & Auguste Genovesio & Yuheng Chen \\
Dingkun Liu & Xinyao Yang & Xinping Xu & Baicheng Chen & Dongrui Wu & Junhao Geng \\
Lexiang Lv & Jianxin Lin & Hanzhe Liang & Jie Zhou & Xuanxin Chen & Jinbao Wang \\
Can Gao & Zhangyi Wang & Zongze Li & Bihan Wen & Yixin Gao & Xiaohan Pan \\
Xin Li & Zhibo Chen & Baorui Peng & Zhongming Chen & Haoran Jin
\end{tabular}
\normalsize 
}
}

\maketitle
\begin{abstract}
\itshape 
This paper reviews the 1st International Workshop on Disentangled Representation Learning for Controllable Generation (DRL4Real), held in conjunction with ICCV 2025. The workshop aimed to bridge the gap between the theoretical promise of Disentangled Representation Learning (DRL) and its application in realistic scenarios, moving beyond synthetic benchmarks. DRL4Real focused on evaluating DRL methods in practical applications such as controllable generation, exploring advancements in model robustness, interpretability, and generalization. The workshop accepted 9 papers covering a broad range of topics, including the integration of novel inductive biases (e.g., language), the application of diffusion models to DRL, 3D-aware disentanglement, and the expansion of DRL into specialized domains like autonomous driving and EEG analysis. This summary details the workshop's objectives, the themes of the accepted papers, and provides an overview of the methodologies proposed by the authors.
\end{abstract}

\section{Introduction}
Disentangled Representation Learning (DRL) is a critical area of research aimed at enabling AI systems to decompose observed data into underlying, interpretable factors of variation. By decoupling complex entities into independent latent factors, DRL holds the potential to address fundamental challenges in AI, such as enhancing the controllability and interpretability of generative systems and improving model generalization.

Despite significant academic interest and progress in DRL methodologies, primarily based on Variational Autoencoders (VAEs) and Generative Adversarial Networks (GANs), the field has largely remained confined to synthetic datasets. The transition to realistic scenarios has been hindered by the complex nature of real-world data and the lack of robust benchmarks and unified evaluation metrics. Traditional DRL methods often struggle when faced with the weaker inductive biases present in real-world environments.

The ICCV 2025 DRL4Real Workshop was organized to address this gap. It aimed to foster the development of novel, realistic datasets and comprehensive benchmarks for evaluating DRL methods in practical applications. The workshop encouraged submissions exploring how DRL can advance model capabilities, with a focus on key areas including controllable generation. This summary provides an overview of the workshop and the contributions of the accepted papers.

\section{Workshop Overview}
The DRL4Real workshop aimed to achieve two primary goals: (1) to provide a comprehensive review of recent developments in applying DRL to realistic scenarios, and (2) to serve as a forum for researchers to explore the challenges and opportunities in controllable generation using disentangled representations.

The workshop attracted diverse submissions spanning various modalities and applications. Following a rigorous review process, 9 papers were accepted for presentation.

\section{Workshop Papers and Themes}
The accepted papers showcased several innovative techniques and highlighted emerging trends in the application of DRL to realistic challenges. We summarize the principal themes observed across the submissions:

\begin{enumerate}
    \item \textbf{DRL for Precise Controllable Generation and Editing.} A major focus was on leveraging DRL to achieve fine-grained control in generation and editing tasks. This included using pre-trained DRL models to extract semantic priors that explicitly constrain edits~\cite{peng2025guided} and modeling spatial reasoning for plausible object placement~\cite{yun2025imagining}.
    \item \textbf{Leveraging Diffusion Models and Novel Inductive Biases.} Recognizing the limitations of traditional regularization, many authors integrated DRL with Diffusion Probabilistic Models (DPMs) and introduced novel inductive biases. This included using textual semantics as a regularization prior~\cite{geng2025textual}, leveraging inherent diffusion properties like time-varying bottlenecks~\cite{gheisari2025divid}, and incorporating structural prompts~\cite{gao2025compress}.
    \item \textbf{3D-Aware and Sequential Disentanglement.} Several papers addressed the challenge of disentangling factors in complex spatial and temporal data. Methods explored 3D-aware generation for autonomous driving~\cite{jin2025controllable}, diffusion-based video factorization~\cite{gheisari2025divid}, reducing static bias in action recognition~\cite{kobayashi2025disentangling}, and semantic isolation theory for 3D anomaly detection~\cite{liang2025fence}.
    \item \textbf{Expanding DRL to Specialized Domains.} The workshop demonstrated the broadening scope of DRL into specialized, realistic domains. Contributions included applications in autonomous driving~\cite{jin2025controllable}, 3D anomaly detection~\cite{liang2025fence}, and EEG analysis for Brain-Computer Interfaces~\cite{chen2025fusiongen}.
    \item \textbf{Foundation Models for Compact Representations.} The interplay between large foundation models and representation learning was also explored, notably in using multimodal LLMs (GPT-4o) to generate high-fidelity images from highly compact textual representations for compression~\cite{gao2025compress}.
\end{enumerate}

\section{Accepted Papers}
This section details the methodologies and contributions of the papers accepted at the DRL4Real workshop, organized thematically.

\subsection{DRL for Controllable Image Generation and Editing}

\subsubsection{A Guided Fine-tuning Framework for Diffusion Models with Disentangled Semantic Priors for Multi-Factor Image Editing~\cite{peng2025guided}}

This paper addresses the challenge of unintended alterations in complex, multi-factor image editing using diffusion models. They propose a Guided Fine-tuning Framework that incorporates disentangled semantic priors as structural constraints.

\begin{figure}[t]
\centering
\includegraphics[width=0.48\textwidth]{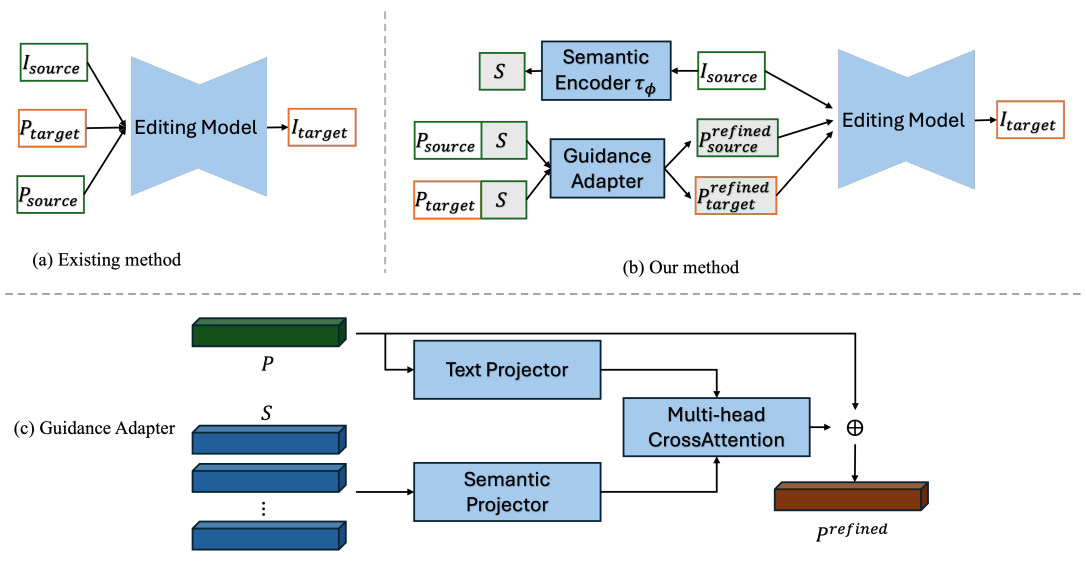}
\caption{Proposed Guided Fine-tuning Framework. (a) Existing methods. (b) Our method using Semantic Encoder and Guidance Adapter. (c) Detailed architecture of the Guidance Adapter. (Source: Paper 9~\cite{peng2025guided})}
\label{fig:guided_finetuning}
\end{figure}

\textbf{Description.} The framework (Figure~\ref{fig:guided_finetuning}) introduces a dual-conditioning approach. While text prompts guide what to change, a disentangled semantic prior guides what to preserve.
\begin{enumerate}
    \item \textbf{Semantic Prior Extraction:} A pre-trained disentanglement model (EncDiff) is used as a Semantic Encoder ($\tau_{\phi}$) to extract a set of independent semantic concept tokens (S) from the input image.
    \item \textbf{Guidance Adapter:} A lightweight, trainable adapter module is integrated into a pre-trained editing model (InfEdit). This adapter fuses the text prompt (P) and the semantic prior (S) using Multi-Head Cross-Attention, where $E_p$ serves as Query and $E_s$ serves as Key/Value.
\end{enumerate}
The resulting structurally-aware prompts ($P^{refined}$) guide the editing model more precisely.

\textbf{Implementation Details.} The framework is validated on the 'DRL for Real' competition Multi-Factor Track dataset. By fine-tuning only the adapter module, the method reduces computational costs. Evaluations using CLIP Score, LPIPS, and FID demonstrated that the framework significantly reduces unintended attribute changes while maximizing desired edits.

\subsubsection{Textual Semantics Matters: Unsupervised Representation Disentanglement in Realistic Scenarios with Language Inductive Bias (TA-Dis)~\cite{geng2025textual}}

This paper proposes Text-Aided Disentanglement (TA-Dis), a framework that leverages the inherently disentangled nature of textual semantics to regularize DRL in the visual domain, addressing the limitations of purely visual constraints in realistic scenarios.

\begin{figure}[H]
\centering
\includegraphics[width=0.48\textwidth]{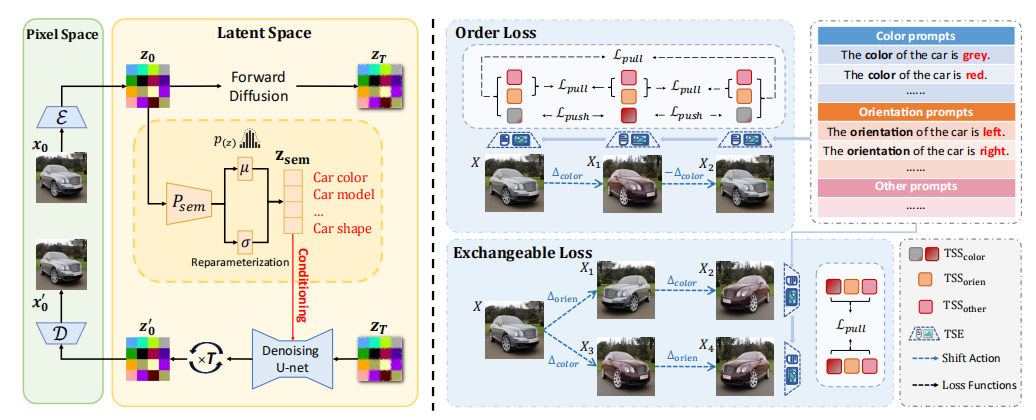}
\caption{Pipeline of the TA-Dis framework. (a) LDM backbone with a semantic projector $P_{sem}$. (b) Text-aided regularization via Order Loss and Exchangeable Loss. (Source: Paper 5~\cite{geng2025textual})}
\label{fig:tadis_pipeline}
\end{figure}

\textbf{Description.} TA-Dis is built upon the Latent Diffusion Model (LDM) in a two-stage approach (Figure~\ref{fig:tadis_pipeline}). This novel use of language as a regularizer builds upon recent trends in unsupervised disentanglement that leverage diffusion model properties~\cite{jin2024closed, yang2024diffusion}, sparse transformations~\cite{song2024unsupervised}, and graph-based reasoning with large language models~\cite{xie2024graph}, all aiming for more robust and meaningful latent representations~\cite{liu2024rate}. The first stage establishes a semantic projector $P_{sem}$ to obtain a semantic code $Z_{sem}$ with primary disentanglement capability. The core innovation is the second stage, introducing Text-Aided Regularization based on CLIP scores to further enhance the disentanglement capability of $Z_{sem}$. Three language inductive biases are designed: Image-Text Alignment (using $\mathcal{L}_{pull}$ and $\mathcal{L}_{push}$), Cycle Consistency (Order Loss $\mathcal{L}_o$), and Shift Equivariance (Exchangeable Loss $\mathcal{L}_e$). These losses regularize the disentanglement process by enforcing constraints in the image-text space.

\textbf{Implementation Details.} The framework utilizes a pretrained VAE and U-Net, optimizing the semantic projector using AdamW. Experiments demonstrated superior disentanglement (measured by TAD) compared to VAE- and other Diffusion-based DRL methods on realistic datasets.

\subsubsection{Imagining the Unseen: Generative Location Modeling for Object Placement~\cite{yun2025imagining}}

This paper tackles the problem of location modeling—determining plausible locations for non-existing objects in a scene. They propose a generative approach to handle the inherent ambiguity and data sparsity of the task.

\begin{figure}[H]
\centering
\includegraphics[width=0.48\textwidth]{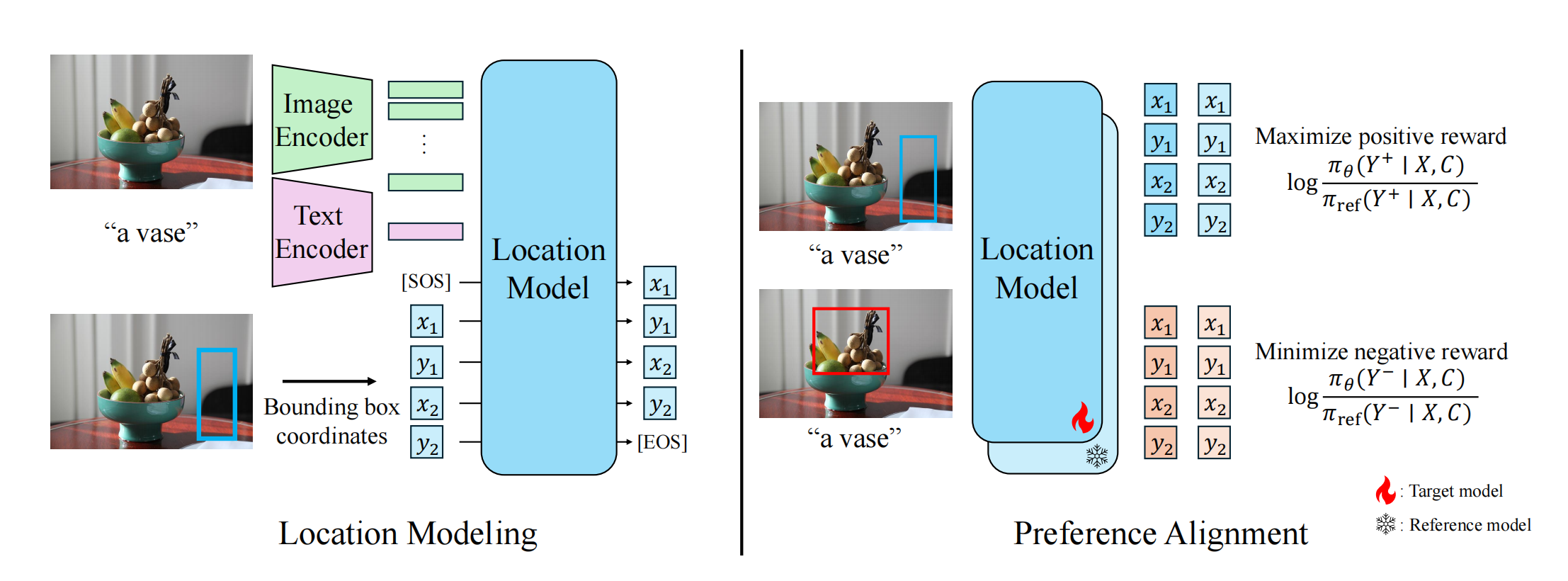}
\caption{Overview of the generative location model. Left: The autoregressive transformer generates bounding box coordinates. Right: Direct Preference Optimization (DPO) refines the model by aligning with positive and negative location preferences. (Source: Paper 1~\cite{yun2025imagining})}
\label{fig:loc_model}
\end{figure}

\textbf{Description.} The authors reframe location modeling as a generative task, $P(Y | X, C)$, using an autoregressive transformer (Figure~\ref{fig:loc_model}, left). The input image and target object class condition the model, which sequentially generates bounding box coordinates. To utilize negative annotations (implausible locations), they incorporate Direct Preference Optimization (DPO), as shown in Figure~\ref{fig:loc_model} (right). DPO fine-tunes the model by maximizing the likelihood that positive locations ($Y^+$) are preferred over negative locations ($Y^-$) based on the Bradley-Terry model.

\textbf{Implementation Details.} The model uses a small GPT-2 architecture, pretrained on the PIPE dataset and fine-tuned on OPA. The generative model achieved superior placement accuracy on OPA and improved visual coherence in object insertion tasks compared to instruction-tuned editing methods.

\subsection{Sequential and 3D Disentanglement}

\subsubsection{DiViD: Disentangled Video Diffusion for Static-Dynamic Factorization~\cite{gheisari2025divid}}

This work introduces \textbf{DiViD}, the first end-to-end video diffusion framework designed explicitly for static-dynamic factorization, aiming to overcome the information leakage common in VAE/GAN approaches.

\begin{figure}[H]
\centering
\includegraphics[width=0.48\textwidth]{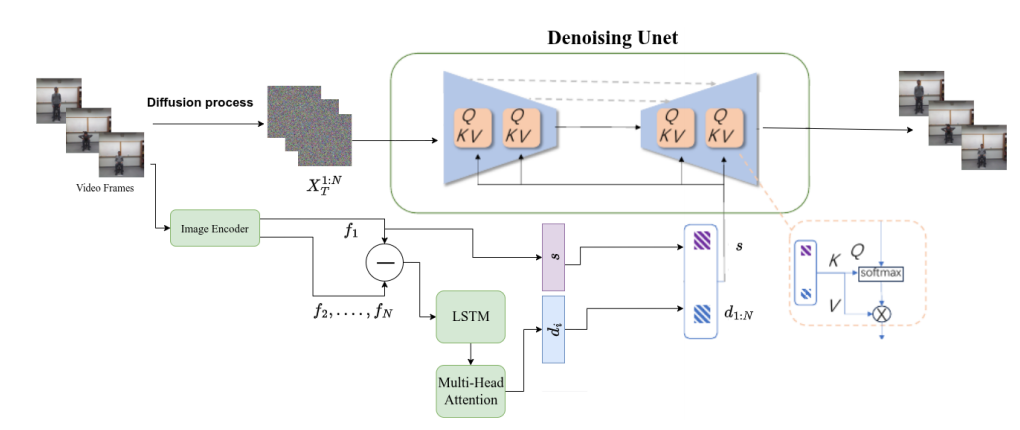}
\caption{Overview of DiViD. The sequence encoder uses residual encoding (subtracting $f_1$) to separate static (s) and dynamic ($d_{1:N}$) tokens, which condition the Denoising U-Net. (Source: Paper 3~\cite{gheisari2025divid})}
\label{fig:divid_overview}
\end{figure}

\textbf{Description.} DiViD incorporates several key inductive biases within a DDPM framework (Figure~\ref{fig:divid_overview}). The sequence encoder employs an \textbf{Architectural Bias} by extracting the static token (s) from the first frame ($f_1$) and dynamic tokens ($d_i$) from the residuals ($f_i - f_1$), explicitly removing static content from the motion code. The decoder utilizes \textbf{Diffusion-driven Inductive Biases}: a Time-Varying Information Bottleneck inherent to the diffusion process, and Cross-Attention Interaction in the U-Net to route global static and local dynamic information appropriately.

\textbf{Implementation Details.} DiViD is trained end-to-end using the standard DDPM loss augmented by an orthogonality regularization term between static and dynamic tokens. Evaluations on MHAD and MEAD showed superior performance compared to SOTA methods.

\subsubsection{Disentangling Static and Dynamic Information for Reducing Static Bias in Action Recognition~\cite{kobayashi2025disentangling}}

This paper addresses the problem of static bias in action recognition, where models rely excessively on static cues rather than dynamic motion.

\begin{figure}[t]
\centering
\includegraphics[width=0.48\textwidth]{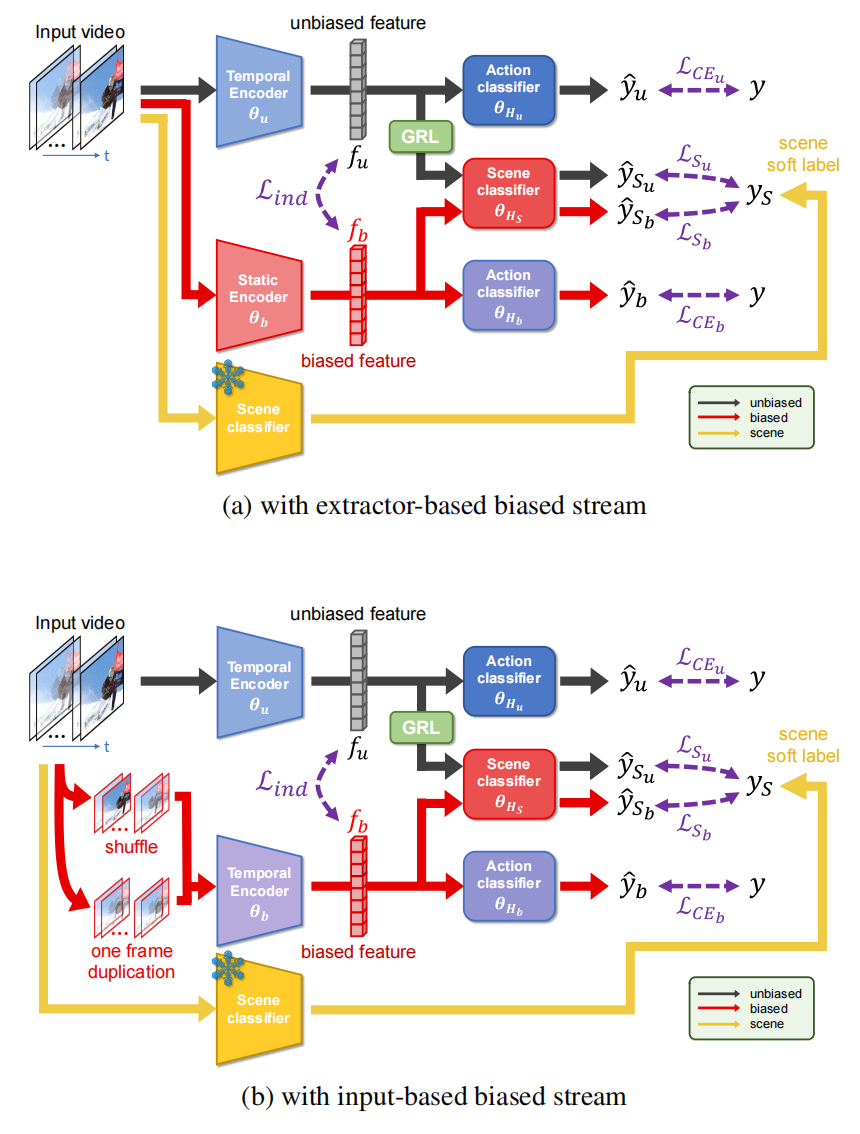}
\caption{The proposed two-stream architecture for disentanglement. An unbiased stream processes temporal dynamics, while a biased stream focuses on static cues. Disentanglement is enforced via an independence loss ($\mathcal{L}_{ind}$) and an adversarial scene prediction loss. (Source: Paper 2~\cite{kobayashi2025disentangling})}
\label{fig:static_dynamic_disentangle}
\end{figure}

\textbf{Description.} A two-stream architecture is proposed to separate unbiased features ($f_u$, dynamics) from biased features ($f_b$, static cues), as illustrated in Figure~\ref{fig:static_dynamic_disentangle}. The biased stream is designed to be inaccessible to temporal information (either via architecture or input manipulation). Disentanglement is achieved through two mechanisms: (1) Statistical Independence Loss ($\mathcal{L}_{ind}$) using the Hilbert-Schmidt Independence Criterion (HSIC) to minimize dependence between $f_u$ and $f_b$. (2) Adversarial Scene Prediction Loss ($\mathcal{L}_S$), using a Gradient Reversal Layer (GRL) on the unbiased stream to force it to fail at scene prediction, thereby removing background information from $f_u$.

\textbf{Implementation Details.} Experiments on datasets emphasizing temporal information demonstrated that the method effectively reduces static bias metrics.

\subsubsection{Controllable Generation with Disentangled Representative Learning of Multiple Perspectives in Autonomous Driving~\cite{jin2025controllable}}

This paper presents a framework for controllable multi-view image generation in autonomous driving scenarios, focusing on disentangling key semantic factors from the scene representation and viewpoint.

\textbf{Description.} The proposed method employs a structured latent representation that decomposes the generative process into three factors: scene content ($z_{sc}$), weather ($z_{w}$), and speed ($z_{s}$). These codes are derived via variational encoding of semantic labels. The architecture utilizes a triplane-based 3D generator conditioned on these latent codes. The triplane representation efficiently maps the latent codes into a compact 3D scene representation. A factor-aware decoder then estimates color and density ($c, \sigma$) for sampled 3D locations, which are synthesized into 2D views via differentiable volumetric rendering (similar to NeRF). This formulation allows independent control over weather, motion, and viewpoint.

\textbf{Implementation Details.} The model is trained with a hybrid objective including reconstruction loss, semantic consistency loss across latent-modified samples, and volumetric rendering-based regularization. The approach was validated on a custom multi-view driving dataset, demonstrating high-fidelity, semantically controllable view synthesis, including view completion.

\subsubsection{Fence Theorem: Towards Dual-Objective Semantic-Structure Isolation in Preprocessing Phase for 3D Anomaly Detection~\cite{liang2025fence}}

This paper addresses the lack of a unified theoretical foundation for preprocessing design in 3D Anomaly Detection (AD). It establishes the \textbf{Fence Theorem} and proposes Patch3D as an implementation.

\begin{figure}[H]
\centering
\includegraphics[width=0.48\textwidth]{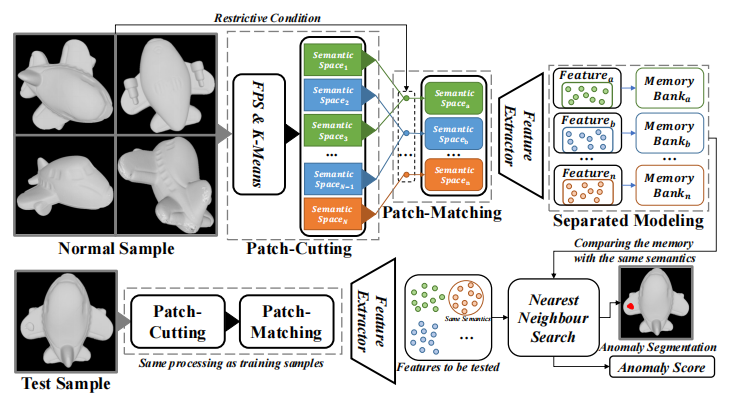}
\caption{Pipeline of Patch3D, illustrating the implementation of the Fence Theorem via Patch-Cutting, Patch-Matching, and Separation Modeling. (Source: Paper 6~\cite{liang2025fence})}
\label{fig:fence_theorem}
\end{figure}

\textbf{Description.} The Fence Theorem formalizes preprocessing as a dual-objective semantic isolator: (1) mitigating cross-semantic interference, and (2) confining anomaly judgments to aligned semantic spaces to establish intra-semantic comparability. The theorem posits that preprocessing aims to divide the point cloud into mutually non-interfering (orthogonal) semantic spaces. Guided by this theorem, the authors implement \textbf{Patch3D} (Figure~\ref{fig:fence_theorem}):
\begin{enumerate}
    \item \textbf{Patch-Cutting:} Uses FPS and K-Means to segment a single point cloud into multiple independent semantic spaces based on its structure.
    \item \textbf{Patch-Matching:} Merges similar semantic spaces across different point clouds to align their meanings.
    \item \textbf{Separation Modeling:} Independently models normal features within each aligned space for anomaly detection.
\end{enumerate}

\textbf{Implementation Details.} The feature extractor utilizes Fast Point Feature Histogram (FPFH) features. Experiments on Anomaly-ShapeNet and Real3D-AD demonstrated that the progressively finer-grained semantic alignment achieved by Patch3D enhances point-level anomaly detection accuracy, validating the Fence Theorem. The proposed theorem and its implementation in Patch3D align with a broader research direction focusing on novel reconstruction techniques~\cite{liang2025taming} and the perception of internal spatial modalities~\cite{Liang_Xie_Hou_Wang_Gao_Wang_2025} to advance the state-of-the-art in 3D anomaly detection.

\subsection{DRL in Specialized Domains and Modalities}

\subsubsection{FusionGen: Feature Fusion-Based Few-Shot EEG Data Generation~\cite{chen2025fusiongen}}

This paper addresses data scarcity in EEG-based BCIs by proposing \textbf{FusionGen}, a data generation framework based on disentangled representation learning and feature fusion for few-shot scenarios.

\begin{figure*}[t] 
\centering
\includegraphics[width=0.9\textwidth]{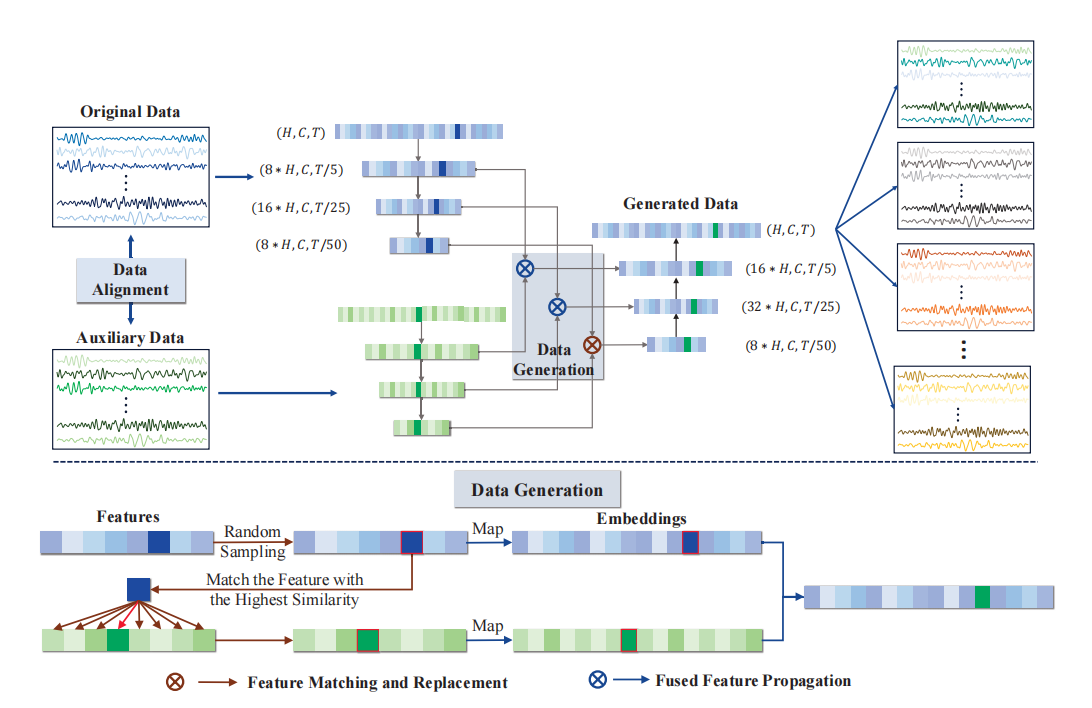}
\caption{Architecture of FusionGen. Aligned raw and auxiliary trials are encoded, fused via the Feature Matching and Replacement mechanism in the latent space, and propagated through the decoder to generate EEG trials. (Source: Paper 4~\cite{chen2025fusiongen})}
\label{fig:fusiongen_arch}
\end{figure*}

\textbf{Description.} FusionGen utilizes a U-Net-shaped encoder-decoder architecture (Figure~\ref{fig:fusiongen_arch}). The core innovation is the \textbf{Feature Matching Fusion} module. This module integrates features across different samples (source and target) in the latent semantic space by randomly sampling target embeddings and replacing them with the most similar source feature. This injects diversity while preserving semantics.

\textbf{Implementation Details.} The network is trained as a denoising autoencoder minimizing MSE. Experiments on three public Motor Imagery (MI) and one steady-state visual evoked (SSVEP) EEG datasets showed that FusionGen significantly outperforms existing augmentation techniques in few-shot scenarios. This approach complements other recent advancements in the EEG domain, such as pipelines for high-quality data construction~\cite{liu2025clean}, foundation models for classification~\cite{liu2025mirepnet}, and techniques for cross-headset distribution alignment~\cite{liu2025spatial}, alongside related work in multi-graph adversarial networks~\cite{liu2025umman}.

\subsection{Generative Models for Compact Representations}

\subsubsection{Why Compress What You Can Generate? When GPT-4o Generation Ushers in Image Compression Fields~\cite{gao2025compress}}

This work explores the potential of large foundation models (GPT-4o) for ultra-low bitrate image compression, advocating for generating pixels rather than compressing them.

\begin{figure}[t]
\centering
\includegraphics[width=0.48\textwidth]{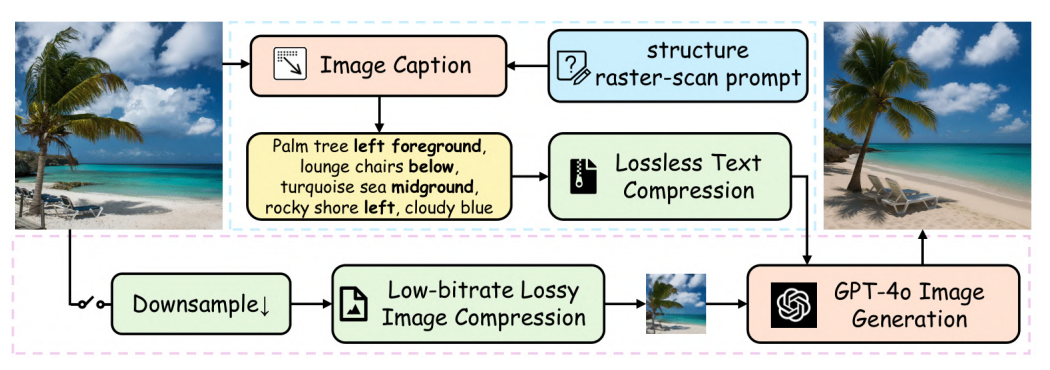}
\caption{Overall pipeline of the multimodal image compression framework based on GPT-4o image generation. (Source: Paper 8~\cite{gao2025compress})}
\label{fig:gpt4o_compression}
\end{figure}

\textbf{Description.} The framework (Figure~\ref{fig:gpt4o_compression}) decouples the image signal into textual descriptions and an optional low-resolution visual prior. The key challenge is maintaining consistency during generation. To address this, the authors introduce \textbf{structural raster-scan prompt engineering}. This method involves designing a prompt that instructs the MLLM (GPT-4o) to describe visual elements in a specific order (top-to-bottom, left-to-right), explicitly preserving the spatial arrangement, object identities, and stylistic properties. The textual description is losslessly compressed and transmitted. If used, the visual condition (downsampled and compressed using MS-ILLM codec) provides base structural information. GPT-4o then reconstructs the image guided by these inputs without any additional fine-tuning.

\textbf{Implementation Details.} Text compression uses Lempel-Ziv coding. Evaluations on the DIV2K validation set showed that the method achieves competitive performance compared to recent generative compression approaches at ultra-low bitrates (e.g., 0.001 bpp).

\section{Conclusion}
The 1st DRL4Real Workshop highlighted significant progress in moving Disentangled Representation Learning from synthetic benchmarks to realistic applications. The accepted papers demonstrated a clear trend towards leveraging powerful generative architectures, particularly diffusion models, and incorporating novel inductive biases, such as language and explicit structural priors, to achieve controllable generation. Furthermore, the expansion of DRL principles into specialized domains like autonomous driving and EEG analysis underscores the growing impact of the field. The innovations presented at the workshop provide valuable insights and pave the way for more robust, interpretable, and controllable AI systems.

\bibliographystyle{IEEEtran}

\begin{thebibliography}{1}

\bibitem{yun2025imagining}
J.~Yun, D.~Abati, M.~Omran, J.~Choo, A.~Habibian, and A.~Wiggers, ``Imagining the unseen: Generative location modeling for object placement,'' in \emph{Proceedings of the IEEE/CVF International Conference on Computer Vision Workshops (ICCVW)}, 2025.

\bibitem{kobayashi2025disentangling}
M.~Kobayashi, N.~Ding, and T.~Tamaki, ``Disentangling static and dynamic information for reducing static bias in action recognition,'' in \emph{Proceedings of the IEEE/CVF International Conference on Computer Vision Workshops (ICCVW)}, 2025.

\bibitem{gheisari2025divid}
M.~Gheisari and A.~Genovesio, ``{DiViD}: Disentangled video diffusion for static--dynamic factorization,'' in \emph{Proceedings of the IEEE/CVF International Conference on Computer Vision Workshops (ICCVW)}, 2025.

\bibitem{chen2025fusiongen}
Y.~Chen, D.~Liu, X.~Yang, X.~Xu, B.~Chen, and D.~Wu, ``{FusionGen}: Feature fusion-based few-shot {EEG} data generation,'' in \emph{Proceedings of the IEEE/CVF International Conference on Computer Vision Workshops (ICCVW)}, 2025.

\bibitem{geng2025textual}
J.~Geng, L.~Lv, and J.~Lin, ``Textual semantics matters: Unsupervised representation disentanglement in realistic scenarios with language inductive bias,'' in \emph{Proceedings of the IEEE/CVF International Conference on Computer Vision Workshops (ICCVW)}, 2025.

\bibitem{liang2025fence}
H.~Liang, J.~Zhou, X.~Chen, J.~Wang, and C.~Gao, ``Fence theorem: Towards dual-objective semantic-structure isolation in preprocessing phase for 3d anomaly detection,'' in \emph{Proceedings of the IEEE/CVF International Conference on Computer Vision Workshops (ICCVW)}, 2025.

\bibitem{gao2025compress}
Y.~Gao, X.~Pan, X.~Li, and Z.~Chen, ``Why compress what you can generate? when {GPT-4o} generation ushers in image compression fields,'' in \emph{Proceedings of the IEEE/CVF International Conference on Computer Vision Workshops (ICCVW)}, 2025.

\bibitem{peng2025guided}
B.~Peng and Z.~Chen, ``A guided fine-tuning framework for diffusion models with disentangled semantic priors for multi-factor image editing,'' in \emph{Proceedings of the IEEE/CVF International Conference on Computer Vision Workshops (ICCVW)}, 2025.

\bibitem{jin2025controllable}
H.~Jin, ``Controllable generation with disentangled representative learning of multiple perspectives in autonomous driving,'' in \emph{Proceedings of the IEEE/CVF International Conference on Computer Vision Workshops (ICCVW)}, 2025.

\bibitem{liu2025spatial}
D.~Liu, S.~Li, Z.~Wang, W.~Li, and D.~Wu, ``Spatial distillation based distribution alignment ({SDDA}) for cross-headset {EEG} classification,'' \emph{arXiv preprint arXiv:2503.05349}, 2025.

\bibitem{liu2025clean}
D.~Liu, Z.~Chen, and D.~Wu, ``{CLEAN-MI}: A scalable and efficient pipeline for constructing high-quality neurodata in motor imagery paradigm,'' \emph{arXiv preprint arXiv:2506.11830}, 2025.

\bibitem{liu2025mirepnet}
D.~Liu, Z.~Chen, J.~Luo, S.~Lian, and D.~Wu, ``{MIRepNet}: A pipeline and foundation model for {EEG}-based motor imagery classification,'' \emph{arXiv preprint arXiv:2507.20254}, 2025.

\bibitem{liu2025umman}
D.~Liu, H.~Zhou, Y.~Qu, H.~Zhang, and Y.~Xu, ``{UMMAN}: Unsupervised multi-graph merge adversarial network for disease prediction based on intestinal flora,'' \emph{IEEE Transactions on Computational Biology and Bioinformatics}, 2025.

\bibitem{jin2024closed}
X.~Jin, B.~Li, B.~Xie, W.~Zhang, J.~Liu, Z.~Li, T.~Yang, and W.~Zeng, ``Closed-loop unsupervised representation disentanglement with $\beta$-vae distillation and diffusion probabilistic feedback,'' in \emph{European Conference on Computer Vision}.\hskip 1em plus 0.5em minus 0.4em\relax Springer, 2024, pp. 270--289.

\bibitem{yang2024diffusion}
T.~Yang, C.~Lan, Y.~Lu, and N.~Zheng, ``Diffusion model with cross attention as an inductive bias for disentanglement,'' \emph{Advances in Neural Information Processing Systems}, vol.~37, pp. 82\,465--82\,492, 2024.

\bibitem{song2024unsupervised}
Y.~Song, T.~A. Keller, Y.~Yue, P.~Perona, and M.~Welling, ``Unsupervised representation learning from sparse transformation analysis,'' \emph{arXiv preprint arXiv:2410.05564}, 2024.

\bibitem{xie2024graph}
B.~Xie, Q.~Chen, Y.~Wang, Z.~Zhang, X.~Jin, and W.~Zeng, ``Graph-based unsupervised disentangled representation learning via multimodal large language models,'' \emph{Advances in Neural Information Processing Systems}, vol.~37, pp. 103\,101--103\,130, 2024.

\bibitem{liu2024rate}
J.~Liu, R.~Feng, Y.~Qi, Q.~Chen, Z.~Chen, W.~Zeng, and X.~Jin, ``Rate-distortion-cognition controllable versatile neural image compression,'' in \emph{European Conference on Computer Vision}.\hskip 1em plus 0.5em minus 0.4em\relax Springer, 2024, pp. 329--348.

\bibitem{Liang_Xie_Hou_Wang_Gao_Wang_2025}
H.~Liang, G.~Xie, C.~Hou, B.~Wang, C.~Gao, and J.~Wang, ``Look inside for more: Internal spatial modality perception for 3d anomaly detection,'' \emph{Proceedings of the AAAI Conference on Artificial Intelligence}, vol.~39, no.~5, pp. 5146--5154, Apr. 2025. [Online]. Available: \url{https://ojs.aaai.org/index.php/AAAI/article/view/32546}

\bibitem{liang2025taming}
H.~Liang, J.~Zhang, T.~Dai, L.~Shen, J.~Wang, and C.~Gao, ``Taming anomalies with down-up sampling networks: Group center preserving reconstruction for 3d anomaly detection,'' \emph{arXiv preprint arXiv:2507.03903}, 2025.

\end{thebibliography}

\end{document}